\pdfoutput=1

\documentclass[11pt]{article}

\usepackage[]{ACL2023}

\usepackage{times}
\usepackage{latexsym}

\usepackage[T1]{fontenc}

\usepackage[utf8]{inputenc}

\usepackage{microtype}

\usepackage{inconsolata}
\usepackage{subfigure}
\usepackage[graphicx]{realboxes}
\usepackage{subcaption}
\usepackage{overpic}
\usepackage{amsmath} 
\usepackage{makecell}
\usepackage{booktabs}
\usepackage{array}

\title{Large Language Model based Situational Dialogues for Second Language Learning}

\author{Shuyao Xu, Long Qin, Tianyang Chen,\\ {\bf Zhenzhou Zha}, \and {\bf Bingxue Qiu}, {\bf Weizhi Wang} \\
Alibaba Group \\
\texttt{\{xushuyao.xsy, ql362507, wangweizhi.wwz\}@alibaba-inc.com} \\ \texttt{\{chentianyang.cty, zhazhenzhou.zzz, mihe.cbx\}@taobao.com}
}

\begin{document}
\maketitle
\begin{abstract}
In second language learning, scenario-based conversation practice is important for language learners to achieve fluency in speaking, but students often lack sufficient opportunities to practice their conversational skills with qualified instructors or native speakers. To bridge this gap, we propose situational dialogue models for students to engage in conversational practice. Our situational dialogue models are fine-tuned on large language models (LLMs), with the aim of combining the engaging nature of an open-ended conversation with the focused practice of scenario-based tasks. Leveraging the generalization capabilities of LLMs, we demonstrate that our situational dialogue models perform effectively not only on training topics but also on topics not encountered during training. This offers a promising solution to support a wide range of conversational topics without extensive manual work. Additionally, research in the field of dialogue systems still lacks reliable automatic evaluation metrics, leading to human evaluation as the gold standard~\citep{smith_etal_2022_human}, which is typically expensive. To address the limitations of existing evaluation methods, we present a novel automatic evaluation method that employs fine-tuned LLMs to efficiently and effectively assess the performance of situational dialogue models. 
\end{abstract}

\section{Introduction}
Research in second language acquisition has emphasized the significance of engaging in relevant exercises within the language being learned \citep{vanpatten2020theories}. However, the shortage of quality language education resources, such as experienced teachers, is a major challenge, especially in some developing countries. Second language acquisition theories suggest that a significant amount of practice is necessary to achieve fluency in speaking \citep{DeKeyser_DeKeyser_2007}. Unfortunately, many students do not have the chance to practice their conversational skills in their target language with a qualified instructor or a native speaker. The significantly improved language learning technologies, powered by advances in natural language processing and AI, such as grammatical error correction \citep{bryant2023grammatical, xu2019erroneous}, automated essay scoring \citep{taghipour2016neural, dong2017attention}, and automatic speech assessment \citep{chen2018end}, have the potential to mitigate the inequity challenge in language learning. However, most of such language learning technologies are designed to help learners improve their vocabulary, grammar, writing, and pronunciation, though conversational skills remain an area where the development of these technologies could further improve.

Educational dialogue systems, particularly those aimed at helping students improve their conversational skills, have been designed as task-oriented dialogue systems  \citep{huang2017chatbot, kwon2018task, li2020developing, Li_2022}. Designing these systems requires substantial effort from experts to define the state and action spaces. Moreover, the interactions they produce can often become tedious and repetitive due to limited variability and personalization~\citep{li2022using}. Consequently, this undermines their effectiveness in engaging learners and hampers their ability to deliver a personalized learning experience.

Due to the outstanding language understanding and generation capacity of large language models (LLMs) \citep{kasneci2023chatgpt, wei2022emergent}, open-ended dialogue systems based on LLMs have been increasingly popular. However, open-ended dialogue systems are not directly suitable for use in the field of language education. A major drawback is their lack of focus on targeted scenario-based practice. When learning a second language, it is beneficial for beginners to start with concrete and practical communication scenarios, which can help them understand and practice the use of the language in specific contexts~\citep{nunan2004task}. Open-ended dialogue systems are typically designed to handle a wide range of topics and types of communication, which means they may not inherently focus on specific scenarios critical for language practice.

To address the issues mentioned above, we propose a topic-based situational dialogue task and situational dialogue models based on LLMs. A situational dialogue task involves interactions confined to specific scenarios or topics, providing a structured yet flexible framework for dialogue. This approach offers a higher degree of freedom compared to task-oriented dialogue, making dialogue more engaging and interesting. Unlike open-ended dialogue, which allows conversation to flow in any direction without topical constraints, situational dialogue confines the interaction to specific topics and focuses language training on these topics to enhance relevance and coherence in those contexts.

The generalization ability of LLMs has been widely noted \citep{radford2019language,brown2020language, chung2022scaling, sanh2021multitask, wei2021finetuned, zhao2023survey}. In the context of situational dialogues, we not only expect models to perform well  on topics covered in the training data (in-domain topics), but we also anticipate that they can generalize to topics not present in the training data (out-of-domain topics), thereby reducing the need for complex manual design. Our experiments demonstrate that our situational dialogue models leveraging the generalization abilities of LLMs have the potential to perform well on out-of-domain topics.

Additionally, we show a baseline that employs a prompt-based approach using ChatGPT~\citep{OpenAI2023ChatGPT} as the dialogue model. Although dialogue systems based on general-purpose LLMs, like GPT-3.5, which usually have hundreds of billions of parameters demonstrate strong performance, the proposed dialogue models based on fine-tuned LLMs with tens of billions of parameters can achieve comparable results.

The evaluation of dialogue models is also a challenge. Although human evaluation methods tend to perform well, they are too costly and slow to be feasible for the rapid iteration of models. On the other hand, automatic evaluation metrics such as BLEU, METEOR and ROUGE are often not reliable enough due to the open-ended nature of conversations \citep{liu2016not}. In this work, we propose novel evaluation metrics, including the response success rate, which measures a dialogue model's ability to generate appropriate individual responses within a dialogue, and the session success rate, which assesses a dialogue model's ability to produce coherent and consistent conversations. Using these metrics, we can assess dialogue interactions at the level of individual utterances as well as the overall conversation. Furthermore, we introduce an automatic evaluation approach employing LLMs that enables efficient and reliable evaluation of situational dialogue models, thereby supporting rapid model optimization.

Overall, our contributions in this work are as follows: 
\begin{itemize}
\item We introduce a situational dialogue approach based on LLMs for second language learning, and release a situational dialogue dataset. Our experiments show that an LLM with 14 billion parameters, fine-tuned on the dataset, can achieve  good performance in the situational dialogue task, outperforming a strong baseline based on GPT-3.5, with the advantage of lower computational costs.
\item Through rigorous testing on diverse unseen topics, we show that our proposed situational dialogue models possess strong generalization capabilities. This is a significant advantage over traditional educational dialogue systems, as we can easily broaden the scope of situational topics without requiring extensive manual effort.
\item We present a novel automatic evaluation method that can evaluate situational dialogue models efficiently and reliably for the situational dialogue task, thereby facilitating convenient and rapid development of situational dialogue models.
\end{itemize}

\begin{figure*}[t]
\centering
\includegraphics[width=13cm]{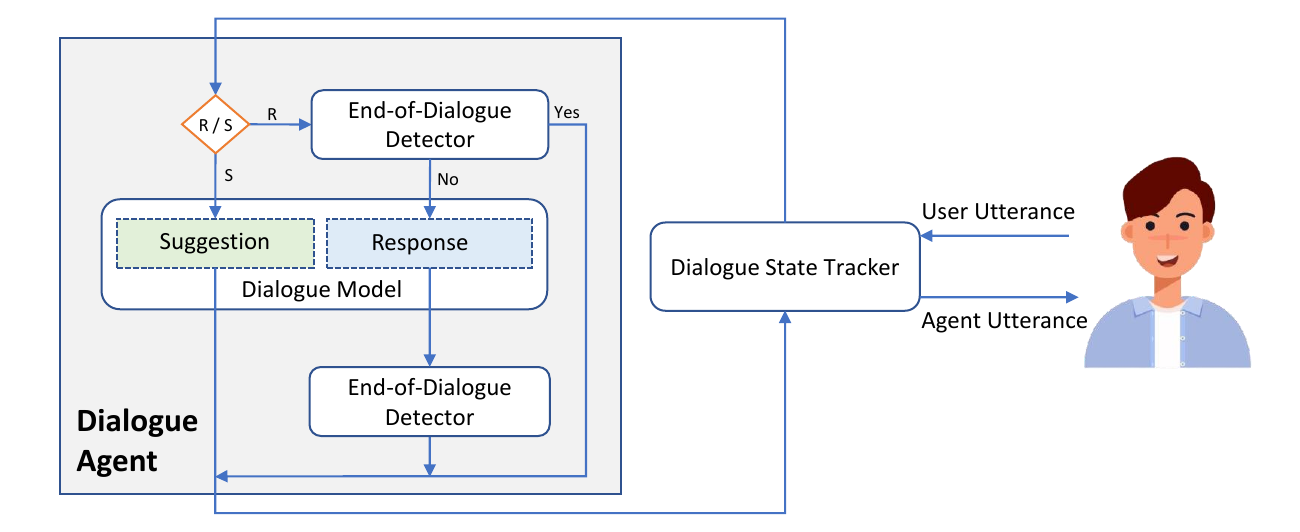}
\caption{Overview of the Dialogue System. The dialogue agent consists of a dialogue model which generates responses and suggestions and an end-of-dialogue detector. In the dialogue agent, "R" and "S" refer to the response generation branch and the  suggestion generation branch respectively. In the response generation branch, the end-of-dialogue detector is applied twice: first, to determine whether the user input concludes the conversation, and second, to assess if the model's response has ended the dialogue.}
\label{fig:dialogue_agent}
\end{figure*}

\section{Related Work}

 \citet{Li_2022} propose a dialogue system for language learning built on the Gunrock 2.0 framework \citep{liang2020gunrock}. This system employs a task-oriented approach that typically includes modules for natural language understanding, dialogue management, and natural language generation. The system treats each topic as a separate task, with each one being a finite state machine, in which the transitions and states are typically predefined, meaning that the dialogue follows a set of predefined rules and paths that determine the conversation flow. Another approach for chatbots designed for language learning is the use of rule-based systems \citep{EHSANI2000167}, where the conversation flow can be directed into different paths based on the user's selections from a predetermined list of utterances. The main limitations of these systems include the restricted flexibility in conversation and the substantial amount of effort needed from experts to develop them. Different from the work mentioned above, the proposed system employs LLMs as a core component of the dialogue system, which can  facilitate  conversations on specific topics while simultaneously offering users considerable flexibility. 

Using automatic evaluation metrics can speed up the progress of conversational system development, but the method for automatically evaluating dialogue models still remains an open question. The commonly used automatic metrics for other natural language generation tasks, like BLEU, have been shown to have a weak correlation with human judgment \citep{liu2016not}. Therefore, human evaluation remains the gold standard in the development of dialogue models \citep{zhu_etal_2023_fireball, lee_etal_2023_prompted}. Unfortunately, how to conduct human evaluation is also an open problem \citep{smith_etal_2022_human}. The two commonly used approaches, which are the single-turn pairwise evaluation and the multi-turn Likert scale assessment, are both flawed \citep{li2019acute}. A recent emerging trend involves using trained metrics: \citet{lowe2017towards} suggest training a model to imitate human judgment when evaluating dialogue responses. \citet{ghandeharioun2019approximating}  suggest an automated evaluation approach with trained scoring models for open-domain dialogue systems, where the dialogue system engages in conversation with itself. However, these systems may not generalize well to data that is different from the data on which they were trained \citep{smith_etal_2022_human}. In contrast to open-ended dialogue, in the context of a situational conversation task, the dialogue needs to revolve around the required topic of conversation. Based on this characteristic, the evaluation of situational dialogue models needs to be tested across all target topics, which often requires the model to handle topics that fall outside the scope of its training data. Therefore, methods that are trained and evaluated solely on in-domain data are not well-suited for the evaluation of situational dialogue tasks. \citet{zheng2023judging} propose to prompt strong LLMs like GPT-4~\citep{openai2023gpt4} to assess the quality of dialogue models' responses on a set of challenging multi-turn open-ended questions, while their approach is intended for evaluating general-purpose chat assistants. Thus, their approach is not well-suited for the situational dialogue task.

\section{Data Generation}
We design 51 situational topics  based on topics commonly covered in English textbooks for Chinese elementary and middle schools, such as hobbies, vacation plans and environmental protection, and corresponding prompts for data generation. The full list of the topics can be found in Table~\ref{in_domain_topics} in Appendix. We use OpenAI's GPT-3.5-turbo\footnote{\url{https://openai.com/chatgpt}} to generate 150 dialogues for each topic,  resulting in a total of 7,650 dialogues. In the process of dialogue generation, we also add instructions to encourage the model to use simple words and sentences, as our target users are second language learners.  Subsequently, we manually refine the generated dialogue data\footnote{The human annotators are the authors of this paper.}; for example, we filter dialogues that contain errors against basic factual knowledge or logical inconsistencies. This process resulted in approximately 3,000 high-quality dialogue entries, selected based on criteria such as relevance to the topic, linguistic accuracy, and simplicity of language suitable for second language learners. We present a prompt example for data generation and an example of generated dialogue in Figure~\ref{fig:prompt_example} and Figure~\ref{fig:dialogue_example} respectively in Appendix.

We hypothesize that situational dialogue models also have the ability to generalize to topics that they have not been trained on. To assess the model's generalization ability, we devise an additional 20 topics that are significantly different from the aforementioned topics, including the Renaissance, the influence of COVID-19, globalization, and other topics detailed in Table~\ref{out_of_domain_topics} in Appendix.  For these topics, we did not employ an LLM to generate dialogue data; instead, these topics serve exclusively to test the model's performance on out-of-domain content.

\begin{figure*}[ht]
    \centering
    \begin{overpic}[width=16cm]{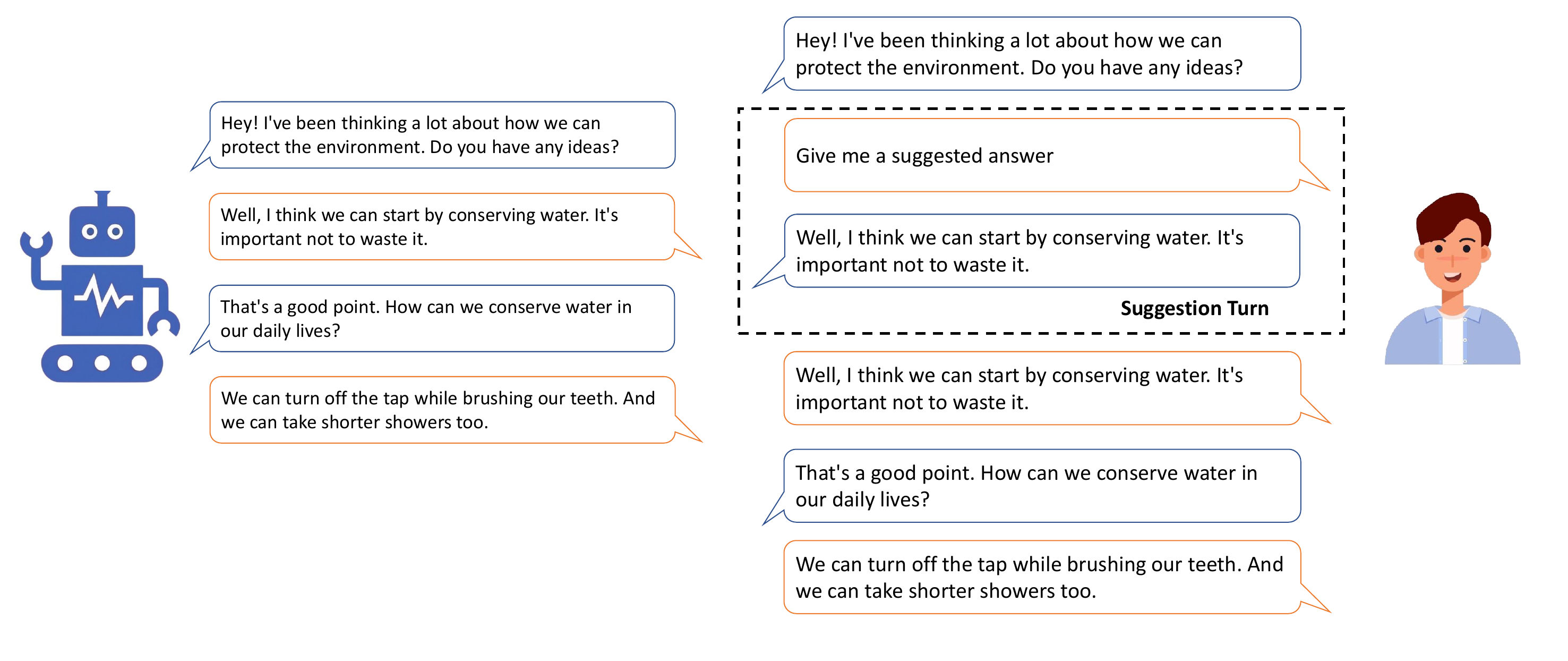}
       \put(12, 0){(a) Orignal Dialogue}
		 \put(47, 0){(b) Dialogue with Suggestion Turn}
    \end{overpic}
    \caption{Suggestion data formulation. Agent utterances are in blue boxes and user utterances are in orange boxes. Suggestion turns are randomly inserted into dialogues for training dialogue models to acquire suggestion generation capability.}
    \label{fig:suggestion-data-formulation}
\end{figure*}

\section{Dialogue Models}

As illustrated in Figure~\ref{fig:dialogue_agent}, our system is composed of a dialogue agent paired with a dialogue state tracker. The dialogue state tracker updates the conversation history with the user's latest utterance, and then passes the updated dialogue state, including the agent's responses, user replies, the topic and whether the conversation has terminated, to the dialogue agent.  If a user requests a suggestion on his/her behalf, the dialogue state is directly passed to the dialogue model, which in turn produces a suggestion. Once the user has received the suggestion, the dialogue state tracker removes both the suggestion request  and the generated suggestion from the dialogue history to avoid confusing the dialogue agent. Otherwise, the dialogue state is sent to an end-of-dialogue detector which is employed to check whether the user has ended the chat. If the dialogue is still ongoing, the dialogue state is fed into the dialogue model for response generation. The newly formed response, along with the dialogue state, is then re-evaluated by the end-of-dialogue detector to see whether the conversation has reached its end.

\subsection{Response Generation}
We implement dialogue models by fine-tuning LLMs on the situational dialogue dataset.  To preserve the LLMs' ability to follow instructions during fine-tuning, we have additionally included 10\% Alpaca data \footnote{The dataset is CC BY NC 4.0 (allowing only non-commercial use).}~\citep{alpaca} in our training dataset. In our experiments, we utilize the Qwen series as base models, specifically selecting the 1.8-billion-parameter version (Qwen-1.8B-Chat), the 7-billion-parameter version (Qwen-7B-Chat), and the 14-billion-parameter version (Qwen-14B-Chat)\footnote{QwenLM/Qwen is licensed under the Apache License 2.0.}~\citep{bai2023qwen}.

Given a multi-turn conversation:

\begin{align}
C=\{(U_{1},A_{1}),(U_{2},A_{2}),...,(U_{N},A_{N})\}, 
\end{align}
where $U$ and $A$ represent utterances from a user and a dialogue agent respectively, with $(U_{n},A_{n})$ forming one dialogue turn.
\begin{align}
U_{n}&=(x_{1}^{n},x_{2}^{n},...,x_{u_{n}}^{n}),\\
A_{n}&=(y_{1}^{n},y_{2}^{n},...,y_{a_{n}}^{n}),
\end{align}
where $u_{n}$ and $a_{n}$ denote the number of tokens of $U_{n}$ and $A_{n}$, $n={1,2,...,N}$.

Throughout the training process, we optimize the model by employing the vanilla next token prediction loss with the teacher forcing strategy. Specifically, the loss function is as follows:

\begin{align}
\mathcal{L}=\frac{1} {\sum_{n=1}^{N} {a_n}} \sum_{n=1}^{N} \sum_{k=1}^{a_{n}} \log p(\hat{y}=y_{k}^{n}\mid X,\theta),
\end{align}
where,
\begin{align}
X&=\{(U_{1},A_{1}),...,(U_{n-1},A_{n-1}), (U_{n}, A^{\prime}_{n})\},\\
A^{\prime}_{n}&= (y_{1}^{n}, y_{2}^{n},...,y_{k-1}^{n})
\end{align}

\subsection{Suggestion Generation}
Unlike open-ended conversations, situational dialogues, are designed to help students practice speaking in their target language. In these exercises, students might sometimes require the dialogue agent to offer constructive suggestions and support in case they encounter difficulties due to their restricted language skills, such as searching for the right vocabulary or constructing grammatically correct sentences.

In our experiments, we facilitate suggestion generation by fine-tuning on the dialogue training data with randomly inserted suggestion turns as illustrated in Figure~\ref{fig:suggestion-data-formulation}. We incorporate suggestion turns by replacing a selected user's utterance with a predefined instruction that seeks suggestions. The original utterance is considered  as the ideal suggestion and is used as the target output for the dialogue models to predict during the training phase, thereby teaching the model to produce contextually appropriate suggestions on the user's behalf.

\subsection{End-of-Dialogue Detector}
Situational dialogues are often briefer than open-domain conversations because they are typically focused on specific tasks or contexts, and there is a finite amount of relevant information that can be exchanged about a particular topic. To this end, we employ an end-of-dialogue detector, which is designed to determine when a conversation has naturally concluded and no further relevant information can be exchanged.

We formulate the end-of-dialogue detection as a binary classification problem based on dialogue history. We fine-tune an extra Qwen-1.8b-Chat model to serve as the end-of-dialogue detector. We construct 2,000 and 500 dialogues for training and testing respectively from our situational dialogue dataset with balanced positive samples (completed dialogues) and negative samples (incomplete dialogues).  The performance of the detector on the test set, as detailed in Table~\ref{End-of-Dialogue-Detector-performance}, demonstrates its ability to accurately predict the completion of dialogues, achieving an accuracy of $98.7\%$.

\subsection{Prompt-based Dialogue Models}
Building conversational systems through prompts with general LLMs is a popular approach~\citep{Lee_2023}. We use GPT-3.5-turbo (as of Dec. 2023) as baseline and generate responses and suggestions by interacting with the dialogue model through its API. Utterances generated in the role of the agent are categorized as responses, and those generated in the role of the user are treated as suggestions. An illustrative example of a prompt used for calling the model's API is provided in Figure~\ref{fig:prompt_baseline} in Appendix.

\begin{table}
\centering
\scalebox{0.9}{
\begin{tabular}{lllc}
\toprule
  Accuracy & Precision & Recall & F1 \\
\hline
 0.987 & 0.991 & 0.983 & 0.987 \\
\bottomrule
\end{tabular}
}
\caption{\label{End-of-Dialogue-Detector-performance}
End-of-Dialogue Detector performance. We take complete dialogues as positive samples.
}
\end{table}

\section{Evaluation}
\begin{figure*}[ht]
\centering
\includegraphics[width=13cm]{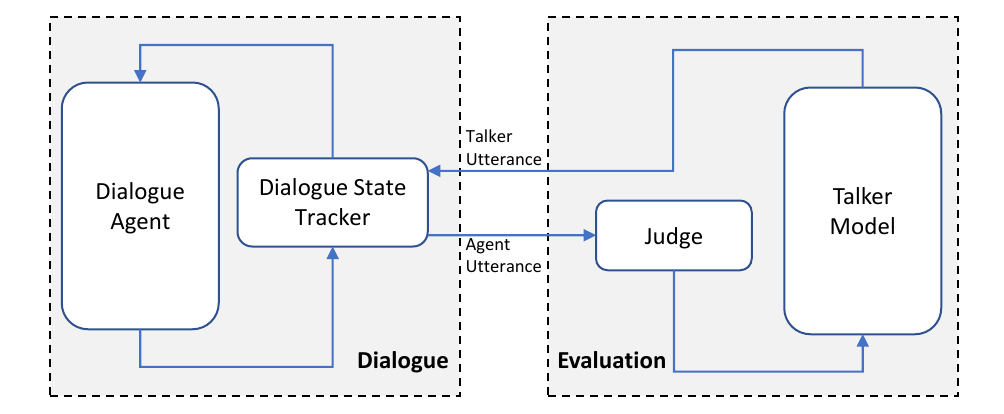}
\caption{Overview of the automatic evaluation pipeline. The evaluation system consists of a judge and a talker model. The judge is employed to assess whether the output of the dialogue agent is an appropriate response to the given context.}
\label{fig:evaluation}
\end{figure*}

\begin{table}
\scalebox{0.9}{
\begin{tabular}{l|lllc}
\toprule
 & Accuracy & Precision & Recall & F1 \\
\hline
Response  & 0.989 & 0.981 & 0.996 & 0.989 \\
Suggestion  & 0.980 & 0.966 & 0.994 & 0.979 \\
\bottomrule
\end{tabular}
}
\caption{\label{judge-performance}
Judge performance on the response testing set and the suggestion testing set. 
}
\end{table}

\subsection{Automatic Metrics}
 For the situational dialogue task, in which each conversation typically consists of several turns of interactions between a dialogue agent and a user, we propose three metrics: response success rate, suggestion success rate, and session success rate. The response success rate, defined as the proportion of correct agent responses to total agent responses, reflects the dialogue model's ability to generate contextually and semantically appropriate agent responses. The suggestion success rate, defined as the proportion of correct suggestions to total suggestions, indicates the model's capability to provide appropriate suggestions on behalf of the user. For automatic evaluation, we train a talker model, as described in Section~\ref{sec:talker}, to simulate a user engaging in a conversation with a dialogue agent. A session refers to a conversation between the dialogue agent and the talker model, which ends either when the conversation naturally concludes (determined by the end-of-dialogue detector), or when the judge detects an incorrect response or an incorrect suggestion from the dialogue agent. A session is deemed successful if it contains no response errors and no suggestion errors throughout the entire conversation. The session success rate, defined as the proportion of successful sessions to the total number of sessions conducted, evaluates the dialogue agent's performance at the conversation level,  providing a  holistic view of the agent's effectiveness in engaging in coherent dialogues.

\subsection{Automatic Evalution Method}
\label{sec:automactic-evaluation-method}

\begin{table}
\scalebox{0.9}{
\begin{tabular}{l|lllc}
\toprule
& \multicolumn{4}{c}{In-domain Topics} \\
 & Accuracy & Precision & Recall & F1 \\
\hline
Response  & 0.971 & 0.977 & 0.993 & 0.985 \\
Suggestion  & 0.977 & 0.985 & 0.991 & 0.988 \\
\midrule
& \multicolumn{4}{c}{Out-of-domain Topics} \\
 & Accuracy & Precision & Recall & F1 \\
\hline
Response & 0.982 & 0.985 & 0.996 & 0.990 \\
Suggestion & 0.970 & 0.972 & 0.996 & 0.984 \\
\bottomrule
\end{tabular}
}
\caption{\label{judge-performance2}
Manual inspection on judge performance on in domain-topics and out-of-domain topics.  
}
\end{table}
We propose an automatic evaluation pipeline as depicted in Figure~\ref{fig:evaluation}. During the evaluation process, a set of initial prompts is provided, where each prompt initiates a conversation on a specific topic. Conversations take place between the dialogue agent and the talker model, with the dialogue state tracker recording the dialogue history. To also evaluate the dialogue agent's ability to generate suggestions, we introduce a random variable, the suggestion rate, within the talker model. This variable determines whether the talker should send a suggestion request to the dialogue agent or generate a user response.

 The judge is used to  assess whether the dialogue agent's responses or suggestions are contextually appropriate for each turn of conversation. The conversation continues until either the end-of-dialogue detector determines that the conversation has concluded, or the judge determines that the dialogue agent has provided a response or suggestion that is either factually incorrect or contextually inappropriate.  

\begin{table}
\centering
\begin{tabular}{l|c}
\toprule
 & Talker Success Rate  \\
\hline
In-domain  & 0.994 \\
Out-of-domain & 0.994 \\
\bottomrule
\end{tabular}
\caption{\label{talker-performance}
Talker performance on the 51 in-domain topics and 20 out-of-domain topics. Talker success rate is defined as a proportion of contextually and semantically appropriate talker responses to total talker responses.
}
\end{table}
\subsubsection{Judge}
\label{sec:Judge}

\begin{table*}
\scalebox{1.0}{
\begin{tabular}{l | p{1.7cm} p{1.7cm} p{1.7cm} | p{1.7cm} p{1.7cm} p{1.7cm}}
\toprule
& \multicolumn{3}{c|}{In-domain Topics} & \multicolumn{3}{c}{Out-of-domain Topics} \\
 & Response  & Suggestion  & Session & Response  & Suggestion  & Session \\
\hline
Qwen-1.8B  & 0.928 & 0.855 & 0.174 & 0.924 & 0.892 & 0.232\\
Qwen-7B & 0.980 & 0.976 & 0.776 & 0.979 & 0.982 & 0.812 \\
Qwen-14B & 0.988 & 0.990 & 0.882  & 0.990 & 0.982 & 0.875 \\
\hline
GPT-3.5-turbo & 0.987 & 0.982 & 0.834  & 0.990 & 0.987 & 0.860 \\
\bottomrule
\end{tabular}
}
\caption{\label{baseline-performance}
Response success rate, Suggestion success rate and Session success rate on 51 in-domain topics and 20 out-of-domain topics. Qwen-1.8B, Qwen-7B and Qwen-14B refer to situational dialogue agents based on Qwen-1.8B-Chat, Qwen-7B-Chat and Qwen-14B-Chat respectively, and GPT-3.5-turbo is the prompt-based dialogue model. 
}
\end{table*}

Dialogue evaluation is conducted at the utterance level by framing the task as a binary classification problem. In this task, a judge is provided with a conversational context and an utterance and is tasked with predicting whether the utterance is an appropriate response by ensuring that it is sensible, coherent, and consistent with the given context.

We divided the situational dialogue dataset by topics, randomly selecting 41 topics  for the training set to cover a diverse range of conversations and reserving  the remaining 10 topics for the testing  to ensure the judge can generalize to new, unseen topics.  From the training topics, we construct 4,000 context and utterance pairs with balanced positive and negative pairs, while from the testing topics, we create 2,000 pairs including a subset of 1,000 response pairs dedicated to assessing the judge performance on responses, and another subset of 1,000 suggestion pairs used to evaluate the judge performance on suggestions. We consider the samples where context and utterance are coherent, sensible and consistent as positive examples. For negative pairs, we randomly select utterances either from different conversations or from the same conversation, provided they are not the ground truth.

We obtain the judge by fine-tuning the Qwen-14B-Chat model  on the aforementioned training pairs. The test results presented in Table~\ref{judge-performance} demonstrate that the judge is capable of making accurate predictions in both response and suggestion scenarios. A higher recall compared to precision shows that the system is biased towards making positive predictions, which is preferable. The negative impact of mistakenly ending a proper conversation is larger than the impact of not ending a conversation that should be terminated. Empirically,  a dialogue model is more prone to generating an incorrect response if the dialogue history already contains an incorrect utterance. Therefore, when the judge mistakenly assesses an incorrect utterance as correct, leading to an erroneous response generated by the dialogue model due to the inclusion of incorrect conversational history, it allows the judge to have more opportunities to correct its previous misjudgments at the conversation level.

To further assess the judge's performance in actual dialogue evaluation, we manually verified approximately 2,000 pair samples, each consisting of a dialogue context and an utterance, randomly selected from a dataset of conversations previously evaluated by the judge, with 1,000 coming from in-domain topics and another 1,000 coming from out-of-domain topics. The results in Table~\ref{judge-performance2} demonstrate that judge exhibits reliable performance in actual dialogue evaluation process, both on in-domain topics and out-of-domain topics. Additionally, we explore the use of GPT-3 prompted with few-shot examples as the judge; however, our experiments did not achieve promising results, and we will conduct further research to investigate potential factors contributing to these outcomes.

\subsubsection{Talker Model}
\label{sec:talker}
We also obtain the talker model by fine-tuning Qwen-14B-Chat. The key distinction in training methods is that the talker model is trained solely for generating user responses, in contrast to the dialogue models, which are optimized primarily for producing agent responses.

We evaluate the performance of the talker model by the judge described in Section~\ref{sec:Judge}. The approach for assessing the talker is similar to the process outlined in Section~\ref{sec:automactic-evaluation-method}, with the difference being that the judge evaluates the talker's utterances and the probability of requesting suggestions is set to 0. In this evaluation, we exclude any sessions that contain erroneous responses from the dialogue agent. As the results presented in Table~\ref{talker-performance}, the talker model is capable of producing reliable responses for both in-domain and out-of-domain topics.

\subsection{Results}
 We evaluate the performance of dialogue agents across 51 in-domain topics. During our experiments, we conduct 30 sessions for each topic, with a suggestion rate of 0.5 in the talker model. As shown in Table~\ref{baseline-performance}, larger models achieve better performance. The Qwen-1.8B based dialogue agent, which performs significantly worse than the Qwen-7B based model and the Qwen-14B based model, is not capable of maintaining coherent and relevant dialogue throughout most sessions without encountering response errors or suggestion errors. In comparison, the Qwen-7B and Qwen-14B based agents can reliably generate reasonable responses and suggestions, resulting in most conversations being error-free. 

We evaluate the generalization ability of dialogue agents on the 20 out-of-domain topics, listed in Table~\ref{out_of_domain_topics} in Appendix. Compared to the results on the in-domain topics, we do not observe a significant performance drop on the out-of-domain topics. Our experiments demonstrate that the situational dialogue models based on Qwen models have the potential to generalize to out-of-domain topics. 

The prompt-based approach based on GPT-3.5-turbo is a strong baseline for the situational dialogue task, achieving better performance than Qwen-7B but weaker than Qwen-14B. However, this GPT-3.5-turbo-based approach may incur higher inference costs, which is a concern in practical applications.

\section{Conclusions}
In this work, we introduce a situational dialogue approach based on fine-tuned LLMs for enhancing the conversational skills of students learning a second language. For evaluation, we employ three metrics: response success rate, suggestion success rate, and session success rate. These metrics allow us to evaluate situational dialogues at both the individual utterance level and the overall conversation level. Additionally, we propose an automatic evaluation method to efficiently and reliably evaluate situational dialogue models. Our experiments demonstrate that fine-tuned 7B or 14B LLMs perform well on the situational dialogue task. Moreover, we show that our proposed situational dialogue models generalize well on topics that do not appear in the training data. Therefore, we can support more topics without requiring extensive manual effort. This is a significant advantage over task-oriented educational dialogue systems~\citep{Li_2022}, which typically require experts to define the dialogue spaces.

\section{Limitations}
The proposed metrics, response success rate, suggestion success rate, and session success rate can only reflect whether a conversation is contextually appropriate, coherent and consistent, but cannot assess the conversations comprehensively, for example, in terms of interestingness~\citep{lee2024evaluating}. In teaching practice, it is essential for conversations to adapt to the learner's language proficiency level. This adaptation should align with the educational content, avoiding using complex vocabulary, phrases or grammatical structures that exceed learners' language capacity. However, we have limited ability to precisely control the language level of content generated by LLMs, and further advancements are needed.

\section{Ethical Statement}
There are concerns regarding the potential toxicity and bias associated with language generation from LLMs. The fine-tuned version of these LLMs in our work may also pose risks of generating offensive or controversial outputs. Considering our target users include a significant number of young students, ensuring a safe deployment and interaction is of utmost importance.

\bibliography{anthology,custom}
\bibliographystyle{acl_natbib}

\appendix
\newpage 
\section{Prompt Templates and Situational Dialogue Topics}
\label{sec:appendix}

\begin{figure*}[t]
\centering
\includegraphics[width=15cm]{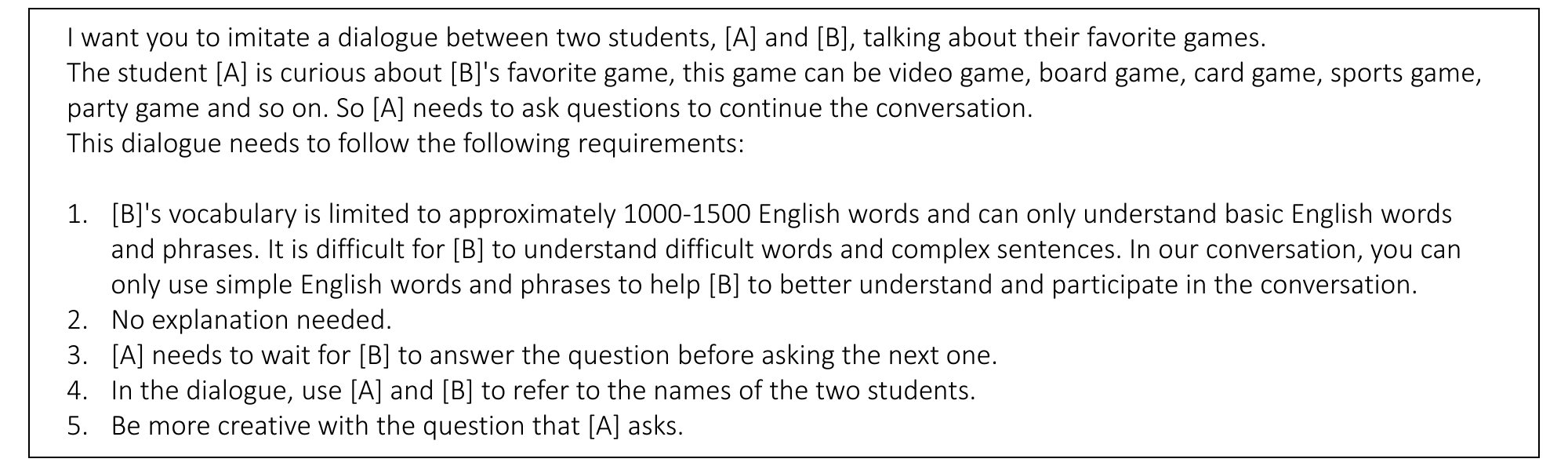}
\caption{A prompt example for generating dialogue.}
\label{fig:prompt_example}
\end{figure*}

\begin{figure*}[t]
\centering
\includegraphics[width=16cm]{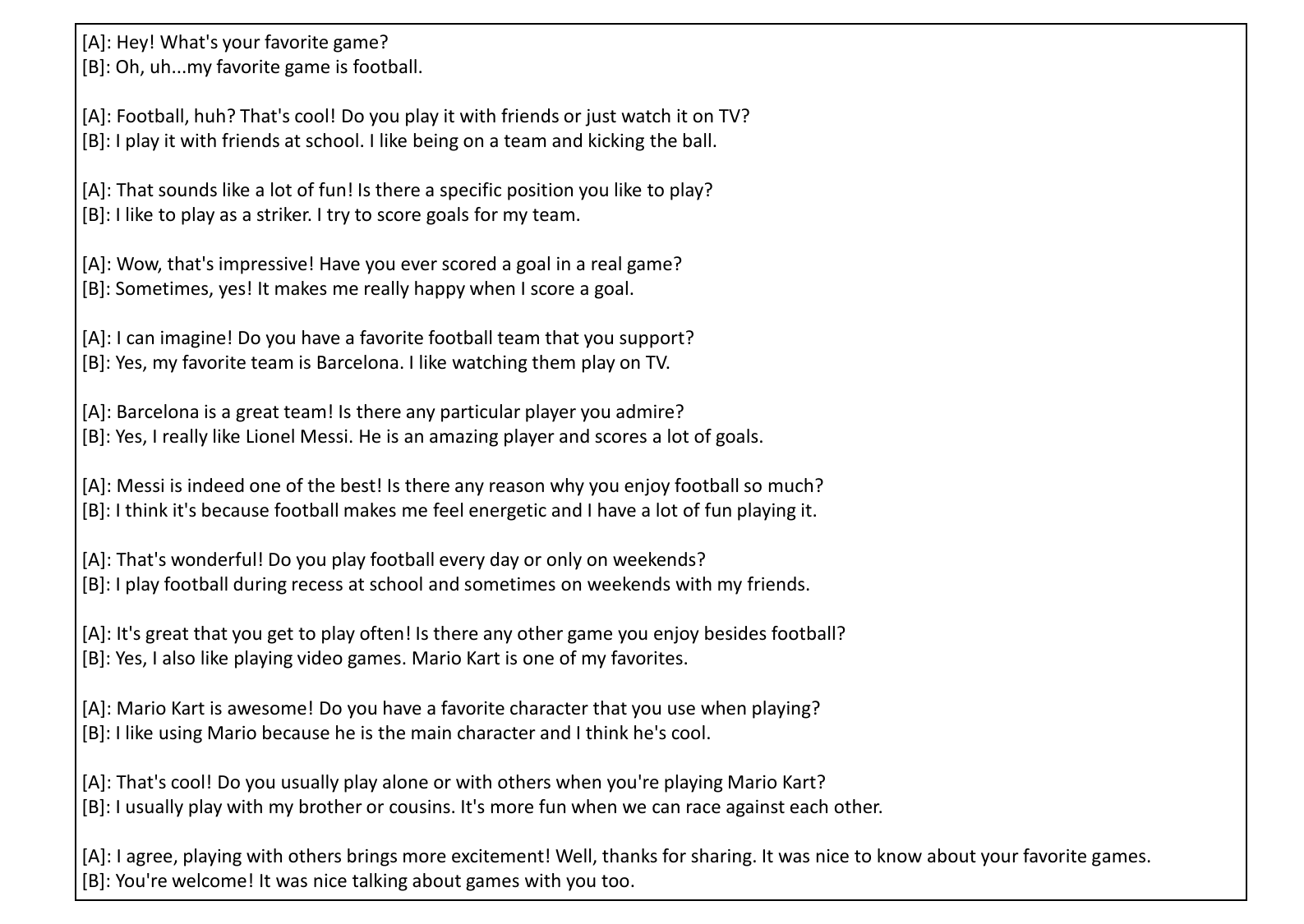}
\caption{An example of generated dialogue.}
\label{fig:dialogue_example}
\end{figure*}

\begin{figure*}[t]
\centering
\includegraphics[width=16cm]{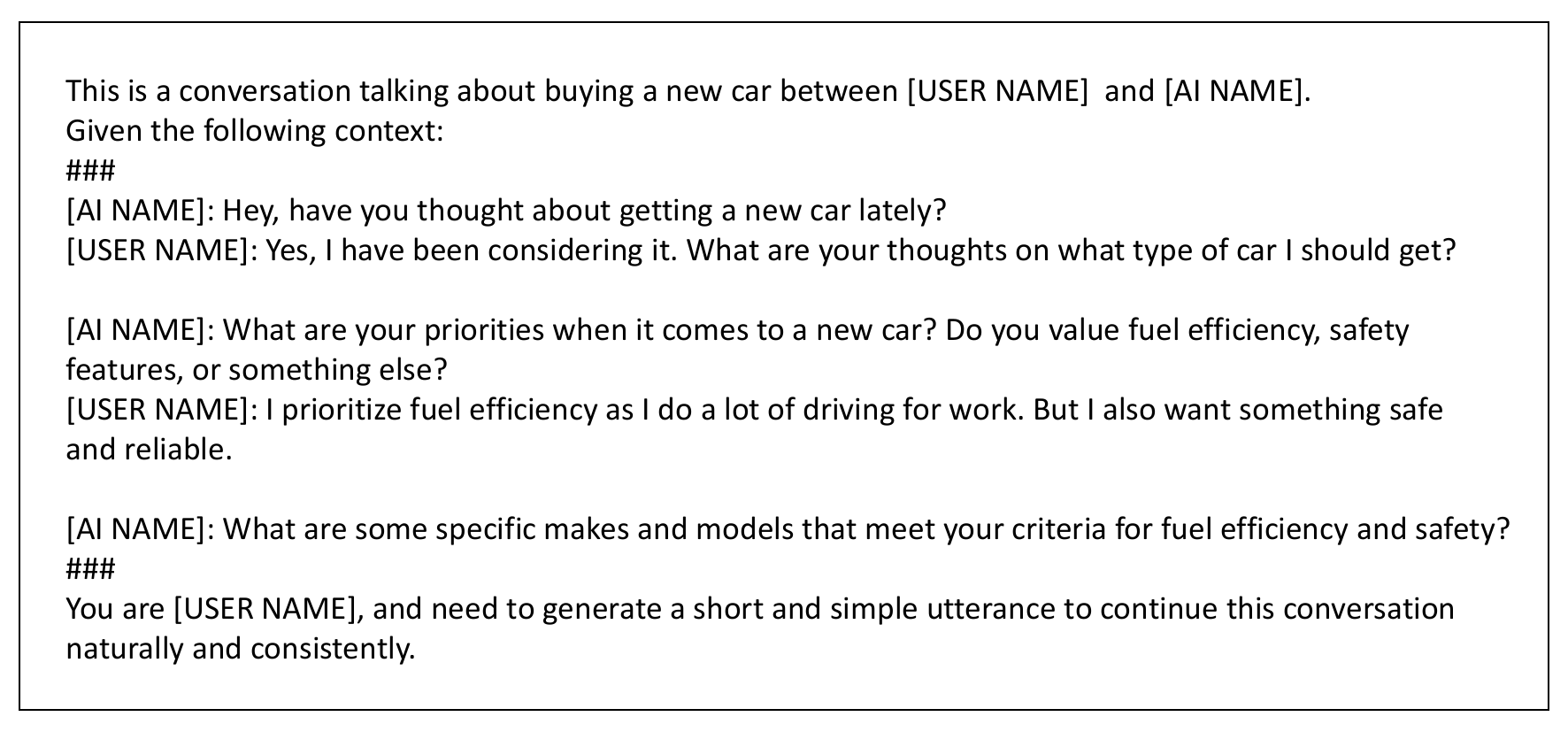}
\caption{A prompt example for prompt-based baseline.}
\label{fig:prompt_baseline}
\end{figure*}

We present a prompt example for dialogue generation in Figure~\ref{fig:prompt_example} and an example of generated dialogue in Figure~\ref{fig:dialogue_example}. An example of a prompt template for the prompt-based dialogue model is shown in Figure~\ref{fig:prompt_baseline}.

The 51 in-domain topics and 20 out-of-domain topics are presented in Table~\ref{in_domain_topics} and Table~\ref{out_of_domain_topics} respectively.

\begin{table*}
\centering
\scalebox{1}{
\begin{tabular}{|p{4cm}|p{4cm}|p{4cm}|}
\hline
 Animals & Best day& Best friend \\
\hline
 Color & Daily routines & Dream house \\
\hline
China & Chinese food & Colleague  \\
\hline
Family & Foreign language & In a restaurant \\
\hline
My hometown & Invention & Job  \\
\hline
Self introduction & Sports & Spring festival  \\
\hline
Travel plan & Vacation plan & Worst weather \\
\hline
Admired person & Advantages and disadvantages & Challenge \\
\hline
Free time & Gift received & Gift for someone   \\
\hline
Scared thing & Seasons & Teacher   \\
\hline
Emotions & Favorite book & Cities \\
\hline
Favorite movie & Environmental protection & My neighbourhood \\
\hline
Job interview & Favorite game & Most impressive thing \\
\hline
 My pets & Superhero & Transportation \\
\hline
Your dreams & Check-in hotel & My birthday \\
\hline
Help & Lie & Choice  \\
\hline
Weather & My hobbies & Free Talk \\
\hline
\end{tabular}
}
\caption{\label{in_domain_topics}
In-domain topics for dialogue generation and model training.
}
\end{table*}

\begin{table*}
\centering
\scalebox{1}{
\begin{tabular}{|p{4cm}|p{4cm}|p{4cm}|}
\hline
 2008 Economic crisis & Drug abuse & Influence of COVID-19 \\
\hline
 Online shopping & Artificial intelligence & Educational equity\\
\hline
Internet addiction & Pros and cons of self driving & Buying a new car  \\
\hline
Extreme sports &  nuclear weapons & The fall of Byzantine empire  \\
\hline
Charity & Gardening & Obesity and keeping a diet \\
\hline
The Renaissance & Climate change & Globalization \\
\hline
Online learning & Traffic safety & \\
\hline
\end{tabular}
}
\caption{\label{out_of_domain_topics}
Out-of-domain topics for evaluating generalizaition ability.
}
\end{table*}

\section{Training Details}
\label{sec:appendix1}

In this work, we utilize the Qwen series\footnote{\url{https://github.com/QwenLM/Qwen}} as our base models. Our training framework builds upon FastChat\footnote{\url{https://github.com/lm-sys/FastChat}}. All models, including the dialogue models, the end-of-dialogue detector, the judge, and the talker model, are trained using the same hyperparameters. We employ a batch size of 128, a learning rate of 2e-5, and a sequence length of 4,096. Each model is trained on its specific training data (dialogue data for dialogue models and the talker model, the end-of-dialogue training set for the end-of-dialogue detector, and the judge training set for the judge model) with an additional 10\% of the Alpaca instruction-following data\footnote{\url{https://github.com/tatsu-lab/stanford_alpaca}}. We train each model for a specific number of training steps: 70 steps for the dialogue models, 140 steps for the talker model, and 210 steps for both the end-of-dialogue detector and the judge. The training is done with 2x A100 GPUs and the longest single training run takes around 12 hours.

\end{document}